%% file: main.tex
\documentclass{article}

\pdfoutput=1

\usepackage[nonatbib,final]{neurips_2020}

\usepackage[utf8]{inputenc} % allow utf-8 input
\usepackage[T1]{fontenc}    % use 8-bit T1 fonts
\usepackage{hyperref}       % hyperlinks
\usepackage{url}            % simple URL typesetting
\usepackage{booktabs}       % professional-quality tables
\usepackage{amsfonts}       % blackboard math symbols
\usepackage{nicefrac}       % compact symbols for 1/2, etc.
\usepackage{microtype}      % microtypography
\usepackage{caption}
\usepackage{subcaption}

\usepackage{times}
\usepackage{mathtools, enumitem}
\usepackage{latexsym}
\usepackage{rel size}
\usepackage{textcomp}
\usepackage{xcolor}
\usepackage{wrapfig}
\usepackage{dsfont}
\usepackage{csvsimple, adjustbox}
\usepackage[font=small,skip=0pt]{caption}
\usepackage[capbesideposition=outside,capbesidesep=quad]{floatrow}
\usepackage{sidecap}
\usepackage{tabularx}
\usepackage{multirow}
\usepackage[flushmargin]{footmisc}

\include{defs}
\newcommand{\ani}[1]{\textcolor{blue}{[Ani: #1}]}
\newcommand\blfootnote[1]{%
  \begingroup
  \renewcommand\thefootnote{}\footnote{#1}%
  \addtocounter{footnote}{-1}%
  \endgroup
}

\title{Deluca -- A Differentiable Control Library: \\
Environments, Methods, and Benchmarking}

\author{
Paula Gradu\textsuperscript{* 1,2}
\And
John Hallman\textsuperscript{* 1}
\And
Daniel Suo\textsuperscript{* 1,2}
\And
Alex Yu\textsuperscript{2}
\And
Naman Agarwal\textsuperscript{2}
\And
Udaya Ghai\textsuperscript{1}
\And
Karan Singh\textsuperscript{1}
\And
Cyril Zhang\textsuperscript{1,3}
\And
Anirudha Majumdar\textsuperscript{1,2}
\And
Elad Hazan\textsuperscript{1,2}
}

\begin{document}

\maketitle

\input{1.abstract}

\input{2.intro}
\input{3.background}

\input{4.experiments}
\input{5.use-cases}

\bibliographystyle{plain}
\bibliography{vent}

\input{appendix.tex}

\end{document}

%% file: defs.tex
\newcommand{\ignore}[1]{}
\DeclareMathAlphabet{\mathbfsf}{\encodingdefault}{\sfdefault}{bx}{n}

\let\Pr\relax
\DeclareMathOperator{\Pr}{\mathbb{P}}

\newcommand{\E}{\mathbb{E}}

\newcommand{\reals}{\mathbb{R}}
\newcommand{\eps}{\varepsilon}

\let\oldtfrac\tfrac
\renewcommand{\tfrac}[2]{\smash{\oldtfrac{#1}{#2}}}

\let\nablaold\nabla
\renewcommand{\nabla}{\nablaold\mkern-2.5mu}

%% file: 1.abstract.tex
\begin{abstract}
We present an open-source library of natively differentiable physics and robotics environments, accompanied by gradient-based control methods and a benchmarking suite. 
The introduced environments allow auto-differentiation through the simulation dynamics, and thereby permit fast training of controllers.  The library features several popular environments, including classical control settings from  OpenAI Gym \cite{brockman2016openai}. We also provide a novel differentiable environment, based on deep neural networks, that simulates medical ventilation.  

We give several use-cases of new scientific results obtained using the library. This includes a medical ventilator simulator and controller, an adaptive control method for time-varying linear dynamical systems, and new gradient-based methods for control of linear dynamical systems with adversarial perturbations. 

\end{abstract}

%% file: 2.intro.tex
\section{Introduction}
\blfootnote{\textsuperscript{*} Equal Contribution \textsuperscript{1} Princeton University \textsuperscript{2} Google AI Princeton \textsuperscript{3} Microsoft Research}  We introduce a new software library for differentiable simulation and control -- Deluca. While there are numerous popular reinforcement learning (RL) libraries, few of them allow for auto-differentiation through the simulation environment. This shortage stems primarily from the intended generality of existing libraries (see below for a brief review). In particular, many popular reinforcement learning settings are inherently discrete (e.g., Atari games \cite{mnih2015human}, board games such as Go \cite{silver2016mastering} and discrete Markov decision processes in general). Furthermore, many robotics and physics-based settings involve the simulation of hybrid dynamical systems (e.g., walking robots or robots manipulating objects). These settings involve inherently non-differentiable dynamics due to discontinuities that occur when an object makes or breaks contact with another object. Existing libraries for reinforcement learning and control typically either attempt to embrace this generality at the expense of differentiability, are proprietary, or do not support common benchmark problems and tools. 

In this manuscript, we attempt to fill these gaps towards an open-source, differentiable simulation and control library. Instead of attempting to address the full spectrum of settings of interest in RL and control, we start from a subset of physics-based environments that constitute some of the most common benchmark problems in RL and continuous control (e.g., differentiable environments for a subset of the OpenAI Gym suite \cite{brockman2016openai}). We also provide a simple API that allows the user to enlarge the scope of environments, for instance, to user-defined deep-learning-based differentiable environments. To enable this, we leverage recent developments in auto-differentiation and accelerated linear algebra in Python as implemented by the Jax library \cite{jax2018github}. We provide several use-cases of our library and demonstrate speed-ups in RL approaches (e.g., policy gradient methods) enabled by access to derivatives. We further propose a benchmarking suite for gradient-based control methods. Our expectation is that the library will enable novel research developments and benchmarking of new classes of RL/control algorithms that benefit from differentiable simulation. 

% \subsection{Related work}
% \label{sec:related work}
\paragraph{Related work.} MuJoCo \cite{todorov2012mujoco}, Bullet \cite{coumans2013bullet} are ubiquitously used as physics engines to benchmark RL algorithms. However, neither provides native auto-differentiation; therefore, derivative-based control algorithms often rely on finite-differences to approximate derivatives. The Drake toolbox \cite{drake} provides high-fidelity simulations of a variety of rigid-body and robotic systems along with implementations of several planning and control algorithms. In contrast, our goal is to provide a lightweight set of differentiable environments that can be used for benchmarking. The DeepMind control suite provides a comprehensive set of control examples and benchmarks \cite{tassa2018deepmind}; our goal is to achieve similar comprehensiveness, but with differentiable environments and methods that exploit this capability. Recently differentiable physics-based simulators have been proposed focusing on end-to-end LCP based rigid body dynamics \cite{de2018end}, deformable objects \cite{hu2019chainqueen} or more general purpose setups such as \cite{degrave2019differentiable,  hu2019difftaichi}. While, these environments offer solutions of varying degree towards differentiable physics, they lack in support towards common benchmark problems and control algorithms.
\vspace{-1em}
% \ka{self note:\begin{enumerate}
%     \item End to End LCP-based rigid body dynamics; released env: cartpole breakout; python; no docs
%     \item Degraves Theano; env: pendulum, biped, robot arm, car; no docs
%     \item CHainQueen focus on deformable objects
%     \item Taichi (ChainQuee++) fluid elastic rigid spring This can do anything, but missing "classical control"
%     \item TinyDiff pendulum, cartpole, quadped
%     \item GradSim parameter est focussed, not available yet
% \end{enumerate}
% words: curated, suitable, familiar, classic, benchmarked planning}

\ignore{
\begin{enumerate}
    \item \ani{There are some control packages in Matlab. One of the most popular is the multi-parametric toolbox 3 (MPT3) from ETH. I wonder if it's worth highlighting that control packages in python are pretty limited. But the growing overlap between the RL and control communities makes a Python package potentially very useful?}
    \item \ani{I think the work on differentiable rendering/vision might also be worth briefly mentioning.}
\end{enumerate}
}

%% file: 3.background.tex
\section{Scientific Background}
\label{sec:background}

\paragraph{Control of dynamical systems:} A discrete-time dynamical system is described as follows: 
$$ x_{t+1} = f_t(x_t, u_t, w_t) .$$
Here  $x_t$ is the state of the system at time-step $t$, $u_t$ is the control input, $w_t$ is the perturbation, and $f_t$ is the transition/dynamics function. We denote sensor observations by $y_t = h_t(x_t)$. Given a dynamical system, the typical problem formulation in RL and optimal control is to minimize the cumulative cost incurred over a finite or infinite time horizon:
% the optimal control problem, also known as reinforcement learning, is to minimize long-term horizon cost given by a sequence of cost functions,
\begin{align*}
& \min_{\pi \in \Pi} J(\pi) = \sum_t c_t(x_t^\pi, u_t^\pi) \quad s.t.\;\; x_{t+1}^\pi = f_t(x_t^\pi, u_t^\pi, w_t).
\end{align*}
Here, $\pi \in \Pi$ is a policy, corresponding to a mapping from (histories of) sensor observations to control inputs. 
%This problem is in general computationally intractable, and theoretical guarantees are available for special cases of dynamics (notably linear dynamics) and perturbations. 
For an in-depth exposition on the subject, see the textbooks by \cite{Bertsekas17,kemin,tedrake,stengel1994optimal}.

\paragraph{The importance of differentiable dynamics} \label{sec:importance}

In many applications, the state space is either continuous/too large to tractably solve the Bellman optimality equation via value iteration or similar methods. 
% Instead, the state space structure has to be taken into account, as well as policies that fit the particular MDP. The most natural algorithm to apply in this situation is policy gradient \cite{sutton2018reinforcement} and its variants. 
Policy gradient methods \cite{sutton2018reinforcement} are among the most popular techniques to tackle such settings. However, the gradient of the rollout cost with respect to the policy naturally involves the dynamics highlighting the importance of differentiable simulators: differentiation through the dynamics allows using deterministic policies leading to sample and computational complexity gains(in contrast to standard policy gradient methods, which employ stochastic policies even for deterministic systems).

The naive way to perform zero-order gradient estimation is via random sampling in policy space, which may be extremely high dimensional (eg. number of weights in a deep neural network). This method involves estimating $J(\pi_{\tilde{\theta}})$ at a parameter $\tilde{\theta}$ close to the current parameters $\theta$ suffering in terms of the variance proportional to the dimension of the parameterization, see e.g. \cite{flaxman2005online,hazan2016introduction}. 

A popular alternative to zero-order gradient estimation is via the likelihood ratio gradient trick (also known as the ``REINFORCE trick'') \cite{aleksandrov1968stochastic,glynn1987likelilood,williams1992simple}, where the gradient computation is reduced as 
$$ \nabla_\theta J( \pi_\theta ) = \E_{\tau} [ \nabla_\theta \log \Pr_\pi(\tau) \cdot J(\tau) ].$$
While this method removes the dependence on the dimension of the parametrization, it is only applicable to stochastic policies, thus introducing a further source of variance to the problem formulation. 

% The variance of a gradient estimator can be reduced from the policy parametrization to something much more managable: the action space. To see this, observe that the classical derivation of the REINFORCE algorithm \cite{sutton2018reinforcement}, the gradient w.r.t. the policy parametrization $\theta$, can be written as
 
% where $\tau$ is a trajectory sampled from the actions of the current policy and dynamics, and $\Pr_\pi(\tau)$ is the probability of this trajectory materializing. The Markovian structure of the problem implies that 
% $$ \nabla_\theta \log \Pr_\pi(\tau) = \sum_t \nabla_\theta \log \Pr_\theta(x_{t+1}|x_t, \pi) , $$
% where $ \Pr_\theta(x_{t+1}|x_t, \pi)$ is the probability of transitioning state at time $t$ under $\pi$. The variance of the latter sum of vectors is proportional to the number of possible actions as opposed to policy parameters! 

% \paragraph{Variance reduction++.} Differentiable environments allow us the ultimate reduction in gradient variance, as we can apply deterministic gradient descent in policy space! 

\paragraph{Algorithms that benefit from differentiable dynamics}
\label{subsec:methods-that-benefit}

The family of methods that can benefit most from differentiable environments are gradient-based methods that fall under the broad umbrella of policy gradient methods, eg. PPO \cite{schulman2017proximal} and TRPO \cite{schulman2015trust}.  
% These can gain a factor of action-space dimension speedup in terms of both sample and computational complexity, as detailed before.  
Another family of methods, and in fact the inspiration for this library,  are recent gradient-based adaptive control methods that fall under the framework of {\it non-stochastic control} \cite{agarwal2019online,hazan2020nonstochastic,simchowitz2020improper,gradu2020adaptive,chen2020black}. These new methods are particularly suitable for differentiable control, since they apply gradient (or higher order) methods on the rollout cost. Lastly, typical continuous-control-oriented planning algorithms leverage first(or higher)-order approximations of the dynamics iteratively. This includes methods such as iLQR/iLQG \cite{todorov2005generalized, mayne1966second} and their Model Predictive Control variants \cite{mayne2014model}. 
We now present experiments that demonstrate the advantage of differentiable dynamics to these methods.

\vspace{-5pt}

%% file: 4.experiments.tex
\vspace{-1em}
\section{Speedups from differentiable dynamics}
\begin{figure}[h]
    \centering
    \includegraphics[width=0.9\textwidth]{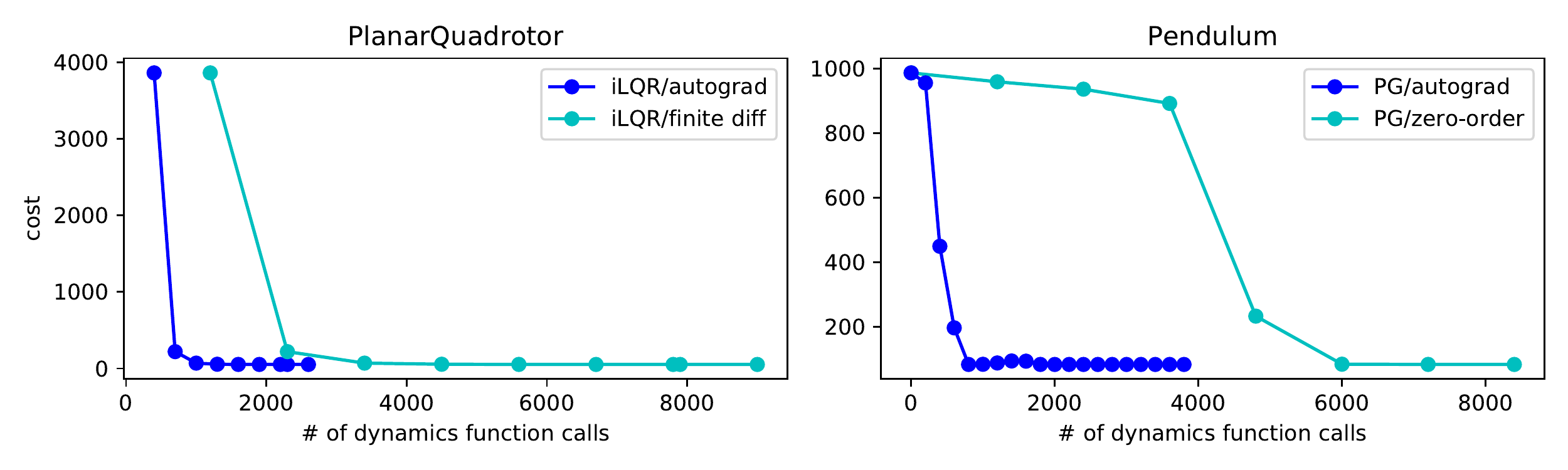}
    \caption{Speedups obtained through differentiable dynamics in two classic settings. \emph{Left:} In \textbf{model-based} planning, linearizations of dynamics can be computed without finite differences. \emph{Right:} In \textbf{model-free} policy optimization, the policy gradient can be evaluated directly rather than by a Monte Carlo estimator.}
    \label{fig:speedups}
\end{figure}

\vspace{-0.5em}
Herein, we present simple experiments on classic environments, showing several aspects of performance gains available via differentiable dynamics and Just-in-Time(JIT) compilation of rollouts.
\vspace{-2em}
\paragraph{Dynamics function oracle complexity.} We quantify the possible algorithmic speedups discussed in Section~\ref{subsec:methods-that-benefit}, in terms of number of calls to the dynamics functions, in two different settings. Results are shown in Figure~\ref{fig:speedups}.
\vspace{-1em}
\paragraph{1) Model-based planning.} We implemented iLQR for trajectory optimization on an underactuated \texttt{PlanarQuadrotor} task (tracking a flight trajectory; $x\in \mathbb{R}^6, u \in \mathbb{R}^2$), using gradients computed by automatic differentiation and finite differences (compute columnwise the Jacobian using $f'(x) \approx [ f(x+\eps)+f(x-\eps) ] / 2\eps$, with $\eps = 10^{-2}$). The solutions were identical (up to gradient approximation error), but the latter required $2\times (\texttt{dim}(x) + \texttt{dim}(u)) = 16$ times more evaluations of the foward dynamics functions per rollout. We could not get iLQR to converge with sampling-based estimators.
\vspace{-1em}
\paragraph{2) Policy gradient.} We considered optimization of a linear policy $\pi \in \reals^{1 \times 2}$ for the \texttt{Pendulum} task, with the conventional ``swing up and stabilize'' reward from OpenAI Gym. To remove confounding factors and uncertainty, we set the initial state to $(\theta = \pi, \dot\theta = 0)$ (rather than uniform at random), and changed the torque constraint from 2 to 10 (to remove the factor of exploration). We used gradient descent(fixed learning rate $\eta = 0.01$) directly on the rollout cost $J(\pi)$ with respect to $\pi$, and compared the deterministic gradient to a Monte Carlo estimator with Gaussian finite differences (known in this setting as  ``evolution strategies''). Aside from requiring fewer dynamics evaluations, the availability of the gradient removes the stochasticity and hyperparameter tuning needed.
\vspace{-2em}
\paragraph{Simulation wall-clock time.} To illustrate the potential for end-to-end performance gains from a purely computational (as opposed to statistical) perspective, we perform a simple comparison of the running times of the \texttt{classic\_control.PendulumEnv} environment provided by Gym, and our replica implementation in Deluca. We timed the initial start-up times, then the simulations over $T = 10^7$ time steps, with all inputs $u_t = 0$ (to isolate the end-to-end computation time of the dynamics only).
The results are shown in Table~\ref{tab:wall-clock-sim}: at the cost of a one-time just-in-time compilation of the dynamics, performed once at the instantiation of the environment, the improvement in the per-iteration time is $>\!1000\times$. This improvement arises from compiler optimizations of functional loops (via \texttt{jax.lax.scan}).

% \begin{SCtable}
% \begin{tabular}{|l|l|l|}
% \hline
%       & startup time & time per iteration \\ \hline
% Gym (NumPy)    & $<1$ $\mu \mathrm{s}$      &  $54.0 \pm 0.2$ $\mu \mathrm{s}$ \\ \hline
% Deluca (Jax) & $203.0 \pm 7.4$ ms & $38.0 \pm 4.2$ ns  \\ \hline
% \end{tabular}
% \caption{Wall-clock time comparison for simulation of the inverted pendulum environment. Just-in-time compilation of the dynamics, while incurring an initial start-up cost, enables \emph{significantly} faster simulations. Standard deviations are computed over $100$ runs.}\label{tab:test}
% \end{SCtable}

% \begin{table}
% \floatbox[\capbeside]{table}
% {\caption{A test table with its caption beside}\label{tab:test}}%
% {\begin{tabular}{ll}
% column 1a & column 2a \\
% column 1b & column 2b \\
% column 1c & column 2c
% \end{tabular}}
% \end{table}
% \begin{table}
% \floatbox[\capbeside]{table}
% {\caption{Wall-clock time comparison for simulation of the inverted pendulum environment. Just-in-time compilation of the dynamics, while incurring an initial start-up cost, enables \emph{significantly} faster simulations. Standard deviations are computed over $100$ runs.}\label{tab:wall-clock-sim}}%
% {
% \begin{tabular}{|l|l|l|}
% \hline
%       & startup time & time per iteration \\ \hline
% Gym (NumPy)    & $<1$ $\mu \mathrm{s}$      &  $54.0 \pm 0.2$ $\mu \mathrm{s}$ \\ \hline
% Deluca (Jax) & $203.0 \pm 7.4$ ms & $38.0 \pm 4.2$ ns  \\ \hline
% \end{tabular}
% }
% \end{table}

\begin{table}[h]
\centering
\begin{tabular}{|l|l|l|}
\hline
      & startup time & time per iteration \\ \hline
Gym (NumPy)    & $<1$ $\mu \mathrm{s}$      &  $54.0 \pm 0.2$ $\mu \mathrm{s}$ \\ \hline
Deluca (Jax) & $203.0 \pm 7.4$ ms & $38.0 \pm 4.2$ ns  \\ \hline
\end{tabular}
\vspace{2mm}
\caption{Wall-clock time comparison for simulation of the inverted pendulum. JIT compilation of dynamics, while incurring start-up cost, enables \emph{significantly} faster simulation. Standard deviations over $100$ runs.}
\label{tab:wall-clock-sim}
\end{table}
\vspace{-2em}

%% file: 5.use-cases.tex
\section{Use cases and examples} \label{sec:use_cases}

\vspace{-0.2cm}
Motivated by the rise of new gradient-based control methods, and in parallel new applications that exploit them, we survey a few recent developments that highlight the need for the library we propose. 

\vspace{-0.2cm}\paragraph{Ventilator control and simulator}
Consider the problem of controlling a medical ventilator for pressure-controlled ventilation. The goal is to control airflow in and out of a patient's lung according to a trajectory of airway pressures specified by a clinician. PID controllers, either hand-tuned or using lung-breath simulators based on gas dynamics, comprise the industry standard for ventilators. In \cite{vent2020}, the authors propose a new data-driven methodology to tackle this problem. First, a deep learning based differentiable simulator was trained on exploratory data, which in turn was used to learn a deep controller. 
Training the controller crucially made use of auto-differentiation. In the absence of a differentiable simulator, accounting for hyperparameter optimization, the pipeline would have taken orders of magnitude more data and compute resources. The figures below are taken from \cite{vent2020}.

\vspace{-0.3cm}\begin{figure}[h]
\RawFloats
\centering
\begin{minipage}{.48\textwidth}
  \centering
  \includegraphics[width=\linewidth]{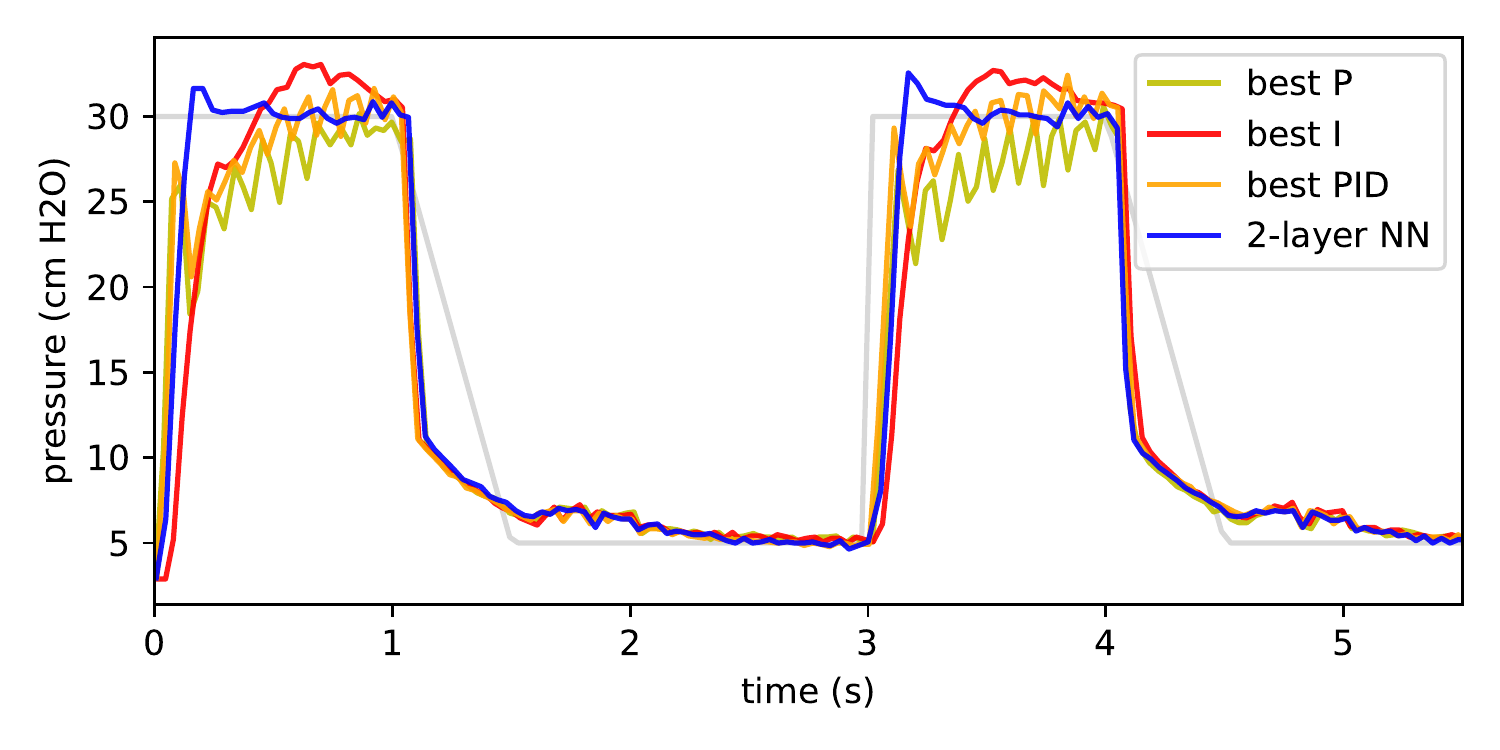}
  \captionof{figure}{Comparison of a learned controller with hand-tuned PID baselines on the ventilation task.}
  \label{fig:ventcontrol}
\end{minipage}
\hspace{.01\linewidth}
\begin{minipage}{.48\textwidth}
  \centering
  \includegraphics[width=\linewidth]{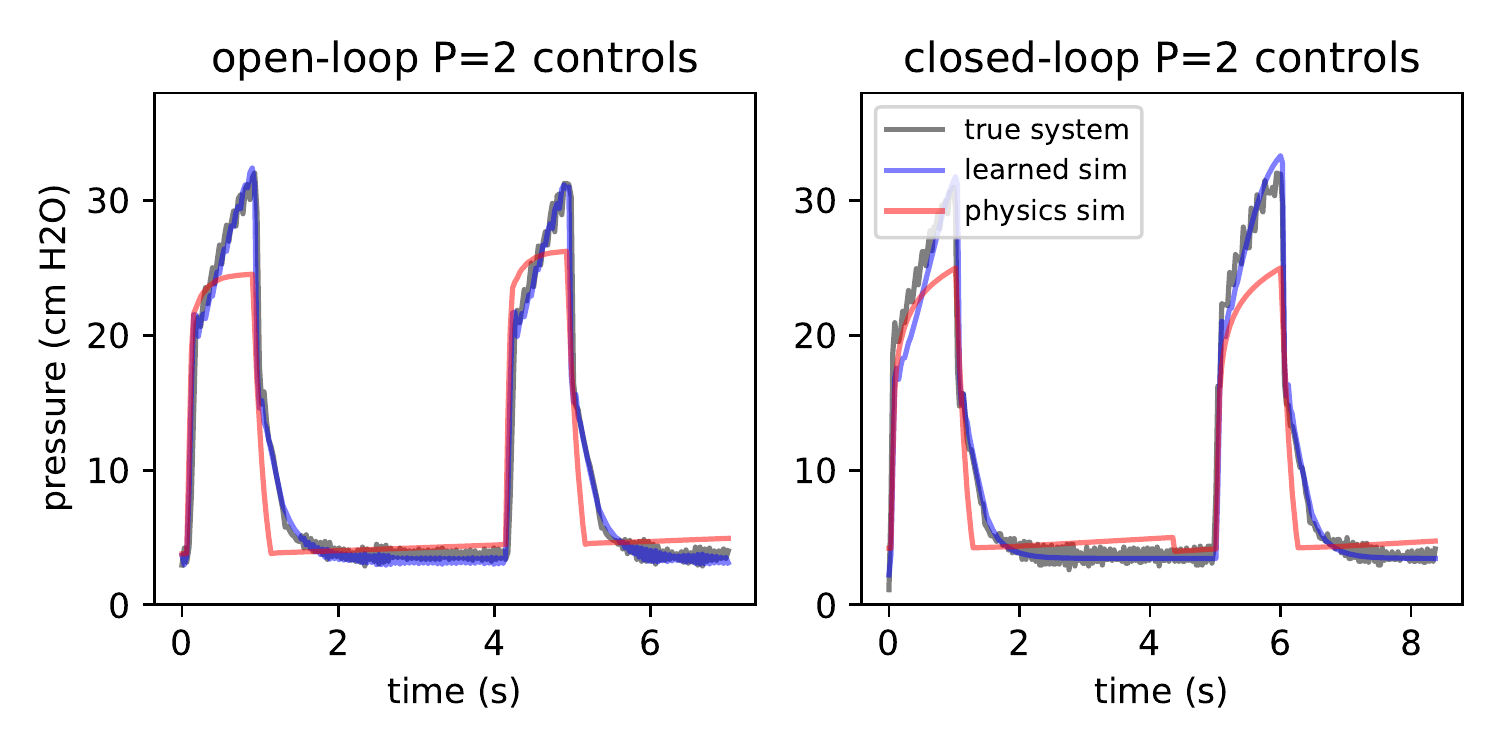}
    \captionof{figure}{Comparison of learned vs. physics-based simulators on open and closed loop controls.}
  \label{fig:ventsim}
\end{minipage}
\end{figure}

\begin{figure}[H]
\centering
\includegraphics[width=.35\linewidth]{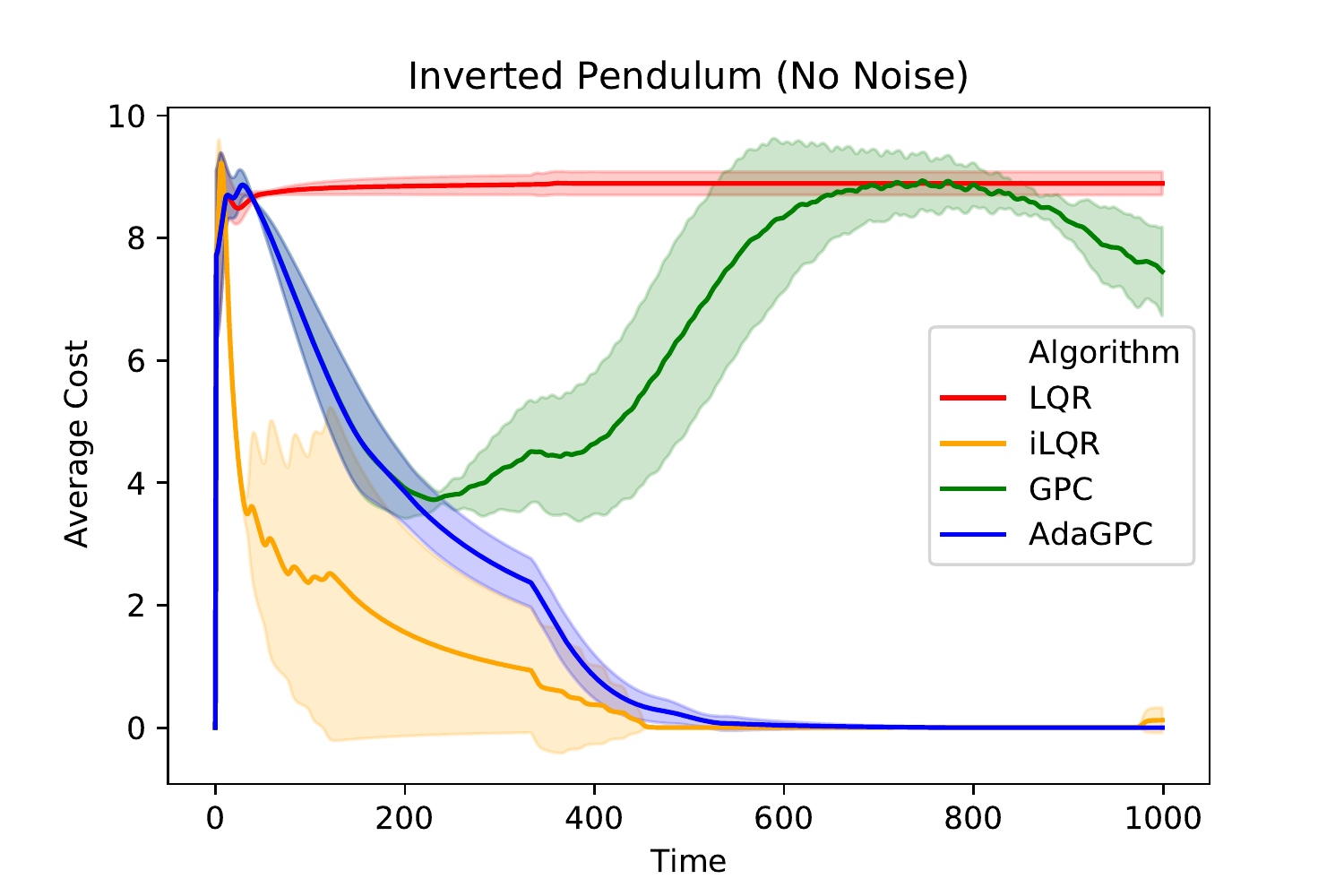}
\includegraphics[width=.35\linewidth]{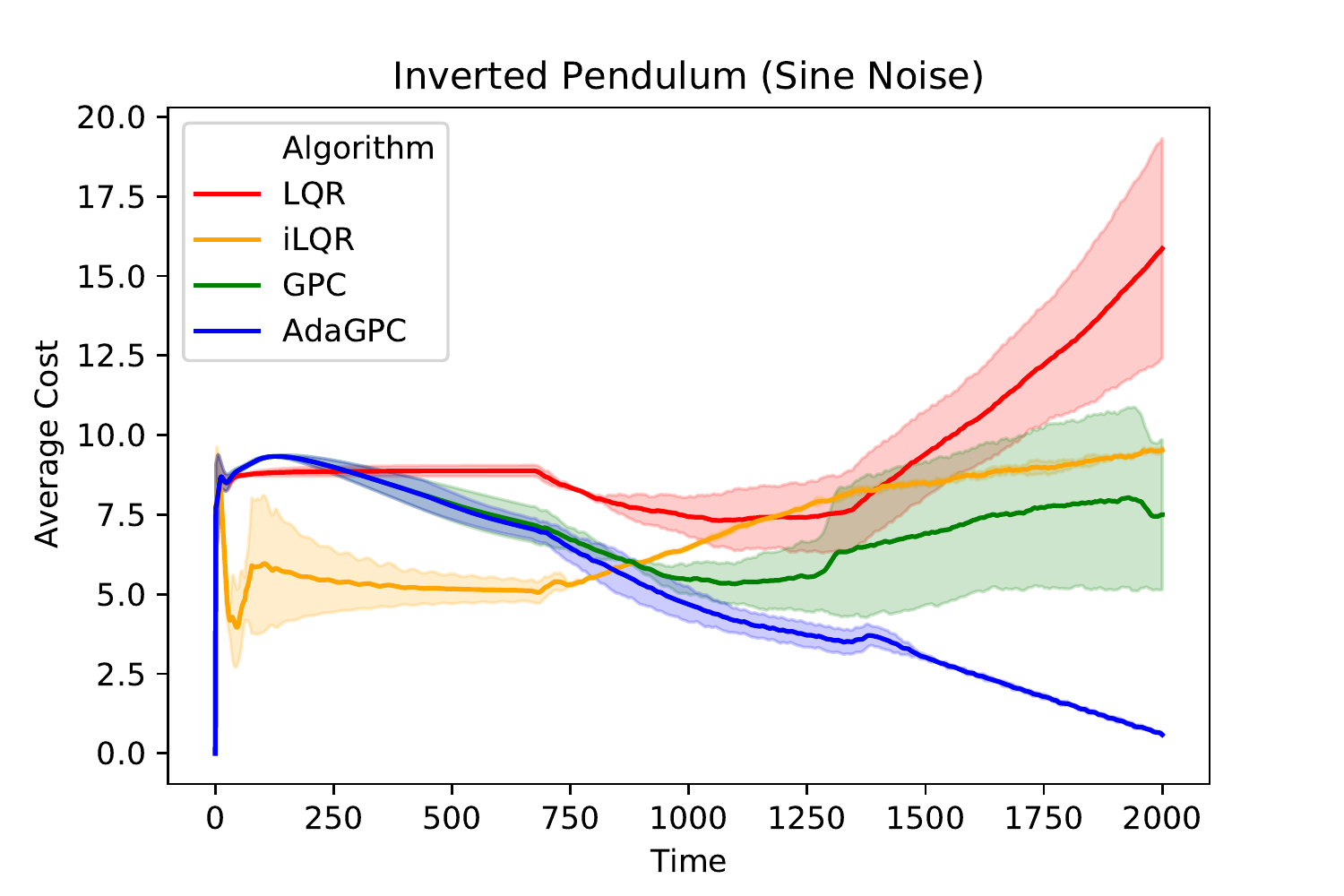}
\caption{LQR, GPC, iLQR, AdaGPC on inverted pendulum -- no noise (left), midway sinusoidal shock (right).}
\label{fig:ada_gpc}
\end{figure}

\vspace{-0.1cm}\paragraph{Adaptive control for the inverted pendulum}
Recent advances in online adaptive control give rise to algorithms with provable guarantees even for time-varying linear dynamical systems \cite{gradu2020adaptive}. These new methods are inherently gradient-based, and differentiate through the dynamics. In figure \ref{fig:ada_gpc}, AdaGPC  is compared against the planning algorithm iLQR. Both algorithms require the dynamics' derivatives so that our library makes their applicability immediate. Furthermore, our library enables experimentation with noise type variations, such as the midway sinusoidal shock in the rightmost plot. We believe studying the trade-offs between planning and online control, as well as the ways to combine the two approaches, is an important new research question that our library helps tackle.
\vspace{-0.3cm}

\vspace{-0.5cm}
\paragraph{Linear-quadratic control with adversarial perturbations}

The inspiration for this library stems from a recent development in control: gradient-based methods that incorporate error feedback into cost functions. The case of linear dynamical systems (LDS) is particularly compelling, as the new methods come with provable guarantees \cite{agarwal2019online,hazan2020nonstochastic,simchowitz2020improper}. Our library comes with these new methods pre-implemented, as well as classical control paradigms such as LQR \cite{kalman1960new} and $\mathcal{H}_\infty$ control \cite{kemin}. %Figure~\ref{fig:noise_plots} profiles different control algorithms in a variety of noise settings.
\vspace{-0.25cm}
\begin{figure}[H]
    \centering
    \includegraphics[width=\linewidth]{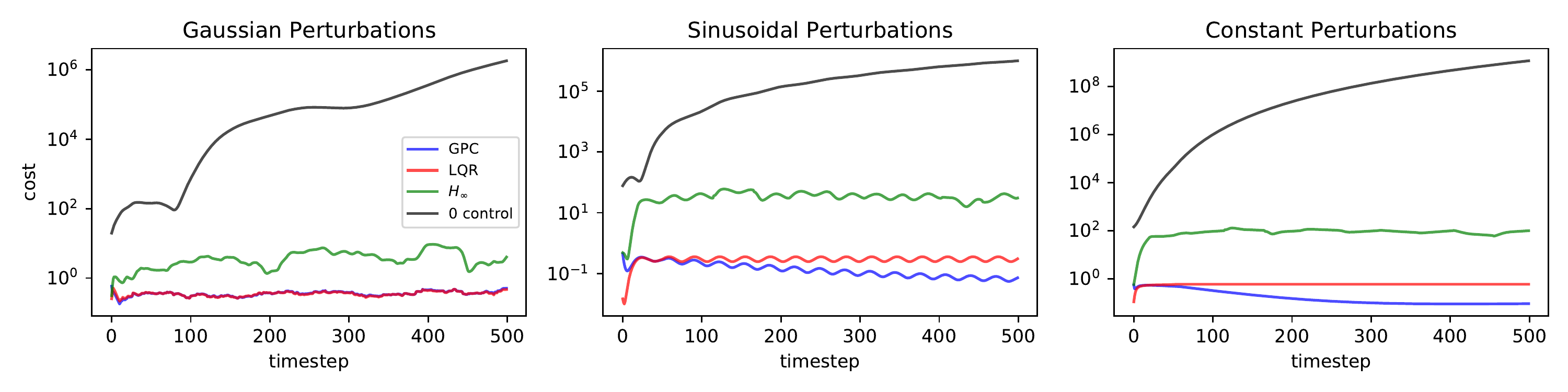}
    \caption{Cost of GPC, LQR and $\mathcal{H}_\infty$ control against Gaussian, sinusoidal, and fixed constant perturbations. }
    \label{fig:noise_plots}
\end{figure}

\begin{ack}

The authors thank Sham Kakade for helpful discussions. 
Part of this work was done when John Hallman, Karan Singh and Cyril Zhang were at Google AI Princeton. Elad Hazan gratefully acknowledges funding from NSF grant \# 1704860. 

\end{ack}

%% file: appendix.tex
\clearpage
\onecolumn
\appendix

\section{\texttt{deluca} library details}

\subsection{Environments}

Below is a table of currently supported environments in \texttt{deluca}. We will continue to add to this list.

\begin{adjustbox}{center}
    \begin{tabular}{|l|l|l|l|l|}
        \hline
         Environment & Obs. Space & Act. Space  & Description  \\
         \hline
         \texttt{classic/acrobot} & \texttt{Box(6,)} & \texttt{Discrete(2)} &  OpenAI Gym Acrobot-v1 \\
         \hline
         \texttt{classic/cartpole} & \texttt{Box(4,)} & \texttt{Discrete(2)} & OpenAI Gym Cartpole-v1 \\
         \hline
         \texttt{classic/mountain\_car} & \texttt{Box(2,)} & \texttt{Box(1,)} & OpenAI Gym \\ & & & MountainCarContinuous-v0 \\
         \hline
         \texttt{classic/pendulum} & \texttt{Box(3,)} & \texttt{Box(1,)} & OpenAI Gym Pendulum-v0 \\
         \hline
         \texttt{classic/planar\_quadrotor} & \texttt{Box(6,)} & \texttt{Box(2,)} & Quadrotor in 2D space\\
         \hline
         \texttt{lds} & \texttt{Box(n,)} & \texttt{Box(m,)} & Linear Dynamical System \\
         \hline
         \texttt{lung/balloon\_lung} & \texttt{Box(1,)} & \texttt{Box(2,)} &  Physics-based lung \\
         \hline
         \texttt{lung/delay\_lung} & \texttt{Box(1,)} & \texttt{Box(2,)} & Physics-based lung \\
         \hline
         \texttt{lung/learned\_lung} & \texttt{Box(1,)} & \texttt{Box(2,)} & Learned lung \\
         \hline
    \end{tabular}
    \label{tab:envs}
\end{adjustbox}

\subsection{Agents}

Below we give a list of the Agents currently available in the library: 

\begin{itemize}
    \item \textbf{Adaptive}: a general adaptive controller that can turn any suitable base controller into its adaptive counterpart.
    \item \textbf{Deep}: a generic fully-connected neural-network controller, to be used as a starting-point for more advanced and specific architectures.
    \item \textbf{DRC}: the recently developed direct response controller \cite{simchowitz2020improper} for partially observable linear dynamical systems.
    \item \textbf{GPC}: the gradient perturbation controller \cite{agarwal2019online, hazan2007logarithmic}.
    \item $\mathbf{H_\infty}$: the $H_\infty$ robust controller \cite{kemin}.
    \item \textbf{iLQR}: the iterative linear quadratic regulator, following the approach described in \cite{tassa2012ilqr}.
    \item \textbf{LQR}: the infinite-horizon, discrete-time linear quadratic regulator \cite{ho1966effective}.
    \item \textbf{PID}: the proportional-integral-derivative action controller \cite{astrom1996pid}.
    \item \textbf{PG}: Controllers trained via deterministic Policy Gradient. 
\end{itemize}

\subsection{Benchmarking}

We currently provide a general-purpose benchmarking tool modeled after \texttt{pytest}'s parametrized fixtures and test functions. This tool allows for parallelized experiments to report results across arbitrary arguments (e.g., environments, agents, parameters). We plan to develop more specific benchmarking suites for various classes of environments and agents.